\documentclass[runningheads]{llncs}
\usepackage[misc]{ifsym}
\usepackage[T1]{fontenc}
\usepackage{ulem}
\usepackage{graphicx}
\usepackage{todonotes}
\usepackage{hyperref}
\definecolor{forestGreen}{rgb}{0.0, 0.54, 0.0}

\definecolor{commentblue}{rgb}{0.0196,    0.4863,    0.8784}

\begin{document}
\title{POV-Surgery: A Dataset for Egocentric Hand and Tool Pose Estimation During Surgical Activities}
\titlerunning{POV-Surgery}
%
\author{Rui Wang\textsuperscript{*}\and
Sophokles Ktistakis\textsuperscript{*}\and
Siwei Zhang\and
Mirko Meboldt\and
Quentin Lohmeyer\textsuperscript{\Letter}}

%
\authorrunning{R. Wang et al.}
\institute{ ETH Zurich, Zurich, Switzerland\\
\email{\{ruiwang46,ktistaks, meboldtm, qlohmeyer\}@ethz.ch}\\
\email{siwei.zhang@inf.ethz.ch}
}
%
\maketitle              
\let\thefootnote\relax\footnotetext{* denotes co-first authorship.}

\begin{abstract}

The surgical usage of Mixed Reality (MR) has received growing attention in areas such as surgical navigation systems, skill assessment, and robot-assisted surgeries. For such applications, pose estimation for hand and surgical instruments from an egocentric perspective is a fundamental task and has been studied extensively in the computer vision field in recent years. However, the development of this field has been impeded by a lack of datasets, especially in the surgical field, where bloody gloves and reflective metallic tools make it hard to obtain 3D pose annotations for hands and objects using conventional methods. To address this issue, we propose POV-Surgery, a large-scale, synthetic, egocentric dataset focusing on pose estimation for hands with different surgical gloves and three orthopedic surgical instruments, namely scalpel, friem, and diskplacer. Our dataset consists of 53 sequences and 88,329 frames, featuring high-resolution RGB-D video streams with activity annotations, accurate 3D and 2D annotations for hand-object pose, and 2D hand-object segmentation masks. We fine-tune the current SOTA methods on POV-Surgery and further show the generalizability when applying to real-life cases with surgical gloves and tools by extensive evaluations. The code and the dataset are publicly available at \url{batfacewayne.github.io/POV_Surgery_io/}.

\end{abstract}

\section{Introduction}
Understanding the movement of surgical instruments and the surgeon's hands is essential in computer-assisted interventions and has various applications, including surgical navigation systems \cite{wesierski2018instrument}, surgical skill assessment \cite{saggio2015objective,goodman2021real,jian2020multitask} and robot-assisted surgeries \cite{fattahi2021mapping}. With the rising interest in using head-mounted Mixed Reality (MR) devices for such applications \cite{palumbo2022mixed,azimi2020interactive,doughty2021surgeonassist,wolf2023different}, estimating the 3D pose of hands and objects from the egocentric perspective becomes more important. However, this is more challenging compared to the third-person viewpoint because of the constant self-occlusion of hands and mutual occlusions between hands and objects. While the use of deep neural networks and attention modules has partly addressed this challenge \cite{liu2021semi,Park_2022_CVPR_HandOccNet,hasson2019learning,lin2021end-to-end}, the lack of egocentric datasets to train such models has hindered progress in this field.
 Most existing datasets that provide 3D  hand or hand-object pose annotations focus on the third-person perspective \cite{zimmermann2019freihand,moon2020interhand2,hampali2020honnotate}. FPHA\cite{garcia2018first} proposed the first egocentric hand-object video dataset by attaching magnetic sensors to hands and objects. However, the attached sensors pollute the RGB frames. More recently, H2O\cite{kwon2021h2o} proposed an egocentric video dataset with hand and object pose annotated with a semi-automatic pipeline, based on 2D hand joint detection and object point cloud refinement.  However, this pipeline is not applicable to the surgical domain because of the large domain gap between the everyday scenarios in \cite{garcia2018first,kwon2021h2o} and surgical scenarios. For instance, surgeons wear surgical gloves that are often covered with blood during the surgery process, which presents great challenges for vision-based hand keypoint detection methods. Moreover, these datasets focus on large,  everyday objects with distinct textures, whereas surgical instruments are often smaller and have featureless, highly reflective metallic surfaces. This results in noisy and incomplete object point clouds when captured with RGB-D cameras. Therefore, in a surgical setting, the annotation approaches proposed in \cite{hampali2020honnotate,kwon2021h2o} are less stable and reliable. Pioneer work in \cite{hein2021towards} introduces a small synthetic dataset with blue surgical gloves and a surgical drill, following the synthetic data generation approach in \cite{hasson2019learning}. However, being a single-image dataset, it ignores the strong temporal context in surgical tasks, which is crucial for accurate and reliable 3D pose estimation~\cite{kwon2021h2o,sener2022assembly101}. Surgical cases have inherent task-specific information and temporal correlations during surgical instrument usage, such as cutting firmly and steadily with a scalpel. Moreover, it lacks diversity, focusing only on one unbloodied blue surgical glove and one instrument, and only provides low-resolution image patches.

To fill this gap, we propose a novel synthetic data generation pipeline that goes beyond single image cases to synthesize realistic temporal sequences of surgical tool manipulation from an egocentric perspective. It features a body motion capture module to model realistic body movement sequences during artificial surgeries and a hand-object manipulation generation module to model the grasp evolution sequences. With the proposed pipeline, we generate POV-Surgery, a large synthetic egocentric video dataset of surgical activities that features surgical gloves in diverse textures (green, white, and blue) with various bloodstain patterns and three metallic tools that are commonly used in orthopedic surgeries. 

In summary, our contributions are:
\begin{itemize}
\item A novel, easy-to-use, and generalizable synthetic data generation pipeline to generate temporally realistic hand-object manipulations during surgical activities.
\item POV-Surgery: the first large-scale dataset with egocentric sequences for hand and surgical instrument pose estimation, with diverse, realistic surgical glove textures, and different metallic tools, annotated with accurate 3D/2D hand-object poses and 2D hand-object segmentation masks.
\item Extensive evaluations of existing state-of-the-art (SOTA) hand pose estimation methods on POV-Surgery, revealing their shortcomings when dealing with the unique challenges in surgical cases from egocentric view.
\item Significantly improved performance for SOTA methods after fine-tuning them on the POV-Surgery training set, on both our synthetic test set and a \textit{real-life test set}.
\end{itemize}

\section{Method}
\begin{figure}[htbp]
    \centering
    \includegraphics[width=\linewidth]{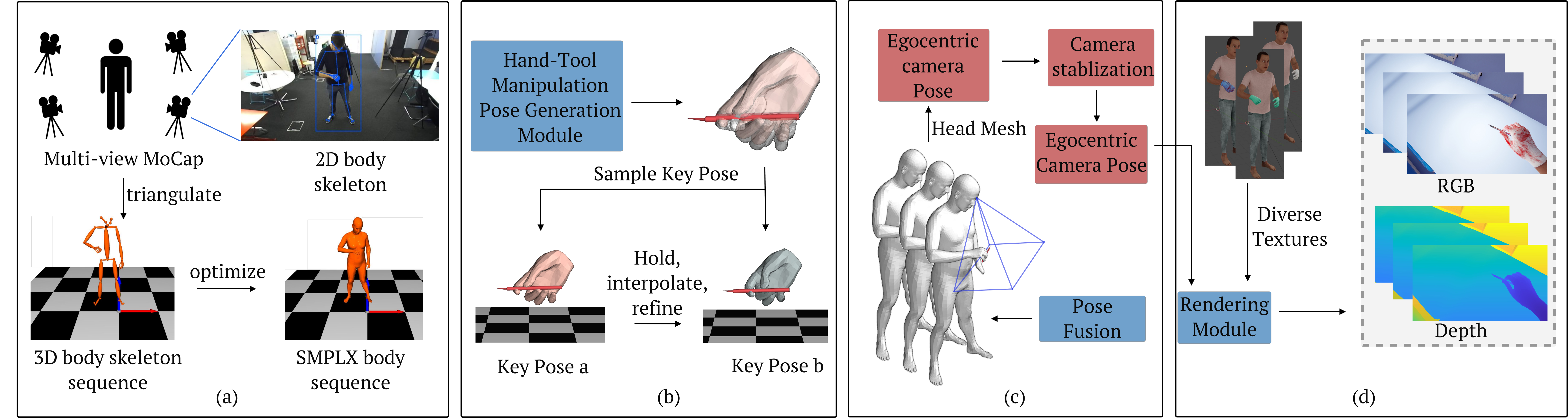}
    \caption{The proposed pipeline to generate synthetic data sequences. (a) shows the multi-stereo-cameras-based body motion capture module. (b) indicates the optimization-based hand-object manipulation sequence generation pipeline. (c) presents the fused hand-body pose and the egocentric camera pose calculation module. (d) highlights the rendering module with which RGB-D sequences are rendered with diverse textures.}
    \label{f:pov_surgery}
\end{figure}

We focus on three tools commonly employed in orthopedic procedures - the scalpel, friem, and diskplacer - each of which requires a unique hand motion. The scalpel requires a side-to-side cutting motion, while the friem uses a quick downward punching motion, similar to using an awl. Finally, the diskplacer requires a screwing motion with the hand. Our pipeline to capture these activities and generate egocentric hand-object manipulation sequences is shown in Fig.~\textbf{\ref{f:pov_surgery}}.

\subsection{Multi-view body motion capture}
To capture body movements during surgery, we used four temporally synchronized ZED stereo cameras on participants during simulated surgeries. The intrinsic camera parameters were provided by ZED SDK and the extrinsic parameters between the four different cameras were calibrated with a chessboard.
We adopt the popular \cite{easymocap}\cite{dong2021fast} module for SMPLX body reconstruction. OpenPose \cite{8765346} with hand - and face-detection modules is first used to detect 2D human skeletons with a confidence threshold of 0.3. The 3D keypoints are obtained via triangulation with camera pose, regularized with bone length. The SMPLX body meshed is optimized by minimizing the 2D re-projection and triangulated 3D skeleton errors. Moreover, we enforce a large smoothness constraint, which regularizes the body and hand pose by constraining the between-frame velocities. It vastly reduces the number of unrealistic body poses.
\subsection{Hand-object manipulation sequence generation}
\begin{figure}[htbp]
    \centering
    \includegraphics[width=1\linewidth]{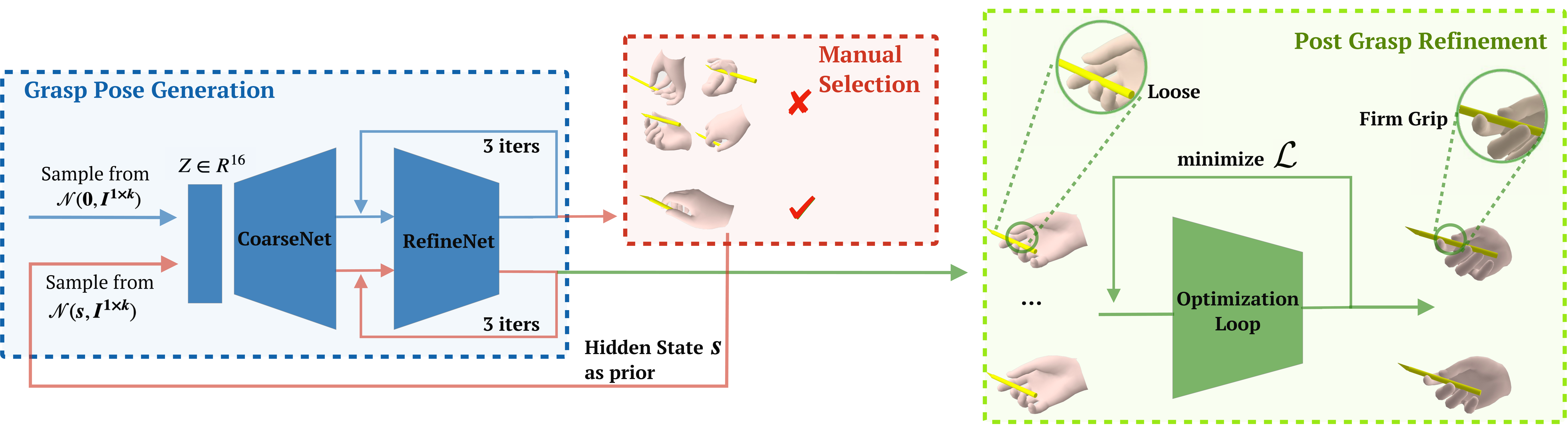}
    \caption{The Hand manipulation sequence generation pipeline consists of three components: grasp pose generation, pose selection, and pose refinement, highlighted in blue, red, and green, respectively. }
    \label{f:pose_gen}
\end{figure}

There are two critical differences between the surgical tool and everyday object grasping: surgical tools require to be held in specific poses. Moreover, a surgeon would hold firmly and steadily with a particular pose for some time span during surgeries. 

To address this issue, we generate each instrument manipulation sequence by firstly modeling the key poses that are surgically plausible, and then interpolating in between to model pose evolution. The key pose generation pipeline is shown in Fig.~\textbf{\ref{f:pose_gen}}.
The part highlighted in blue is the pose generation component based on GrabNet \cite{GRAB:2020}. We provide the 3D instrument models to GrabNet with arbitrary initial rotation and sample from a Gaussian distribution in latent space to obtain diverse poses. 500 samples are generated for scalpel, diskplacer, and friem, respectively, followed by manual selection to get the best grasping poses as templates. With a pose template as prior, we perform the re-sampling near it to obtain diverse and similar hand-grasping poses as key poses for each sequence. To improve the plausibility of the grasping pose and hand-object interactions, inspired by \cite{jiang2021graspTTA}, an optimization module is adopted for post-processing, with the overall loss function defined as:
\begin{equation}
    \mathcal{L} = \alpha \cdot L_{penetr} + \beta \cdot L_{contact} + \gamma \cdot L_{keypoint},
\end{equation}
$ L_{penetr}$, $ L_{contact}$,  and $L_{keypoint}$ denote the penetration loss, contact loss, and keypoint loss, respectively. And $\alpha$, $\beta$, $\gamma$ are object-specific scaling factors to balance the loss components. For example, the weight for penetration is smaller for the scalpel than the friem and diskplacer to account for the smaller object size.
The penetration loss is defined as the overall penetration distance of the object into the hand mesh:
\begin{equation}
L_{penetr }=\frac{1}{\left|\mathcal{P}_{i n}^o\right|} \sum_{p \in \mathcal{P}_{i n}^o} \min _i\left\|p-\mathcal{V}_i\right\|_2^2,
\end{equation}
where $\mathcal{P}_{i n}$ denotes the vertices from the object mesh which are inside the hand mesh,
and $\mathcal{V}_i$ denotes the hand vertex. The $\mathcal{P}_{in}^o$ is defined as the dot product of the vector from the hand mesh vertices to their nearest neighbors on the object mesh. To encourage hand-object contact, a contact loss is defined to minimize the distance from the hand mesh to the object mesh.
\begin{equation}
    L_{contact}=\sum_j{\min _i\left\|\mathcal{V}_i-\mathcal{P}_j\right\|_2^2},
\end{equation}
where $\mathcal{V}$ and $\mathcal{P}$ denote vertices from the hand and object mesh, respectively. In addition, we regularize the optimized hand pose by the keypoint displacement, which penalizes hand keypoints that are far away from the initial hand keypoints:
\begin{equation}
    L_{keypoint} = \sum_i{\left\|K_i - k_i\right\|^2},
\end{equation}
where $K$ is the refined hand keypoint position and $k$ is the source keypoint position. 
After the grasping pose refinement, a small portion of the generated hand poses are still unrealistic due to the poor initialization. To this end, a post-selection technique similar to\cite{tzionas2016capturing,hasson2019learning} is further applied to discard the unrealistic samples with hand-centric interpenetration volume, contact region, and displacement simulation.

For each hand-object manipulation sequence, we select 30 key grasping poses, hold on, and interpolate in between to model pose evolution within the sequence. The number of frames for the transition phase between every two key poses is randomly sampled from 5 to 30. The interpolated hand poses are also optimized via the pose refinement module with the source keypoint in $L_{keypoint}$ defined as the interpolated keypoints between two key poses.

\subsection{Body and hand pose fusion and Camera Pose Calculation}
In previous sections, we individually obtained the body motion and hand-object manipulation sequences. To merge the hand pose into the body pose to create a whole-body grasping sequence, we established an optimization-based approach based on the SMPLX model. The vertices to vertices loss is defined as:
\begin{equation}
L_{V2V }=\sum_{\hat{v}_i \in P_{hand}} \left\|v_{M(i)}-(R\hat{v}_i + T)\right\|_2^2,
\end{equation}
where $\hat{v}$ is the vertices in the target grasping hand, $v$ is the vertices in the SMPLX body model, with $M$ being the vertices map from MANO's right hand to SMPLX body. $R$ and $T$ are the rotation matrix and translation vector applied to the right hand. The right-hand pose of SMPLX, $R$, and $T$ are optimized with the Trust Region Newton Conjugate Gradient method (Trust-NCG) for 300 iterations to obtain an accurate and stable whole-body grasping pose. $R$ and $T$ are then applied to the grasped object. The egocentric camera pose for each frame is calculated with head vertices position and head orientation. Afterwards, outlier removal and moving average filter are applied to the camera pose sequence to remove temporal jitterings between frames.

\subsection{Rendering and POV-Surgery dataset statistics}
\begin{figure}[htbp]
    \centering
    \includegraphics[width=\linewidth]{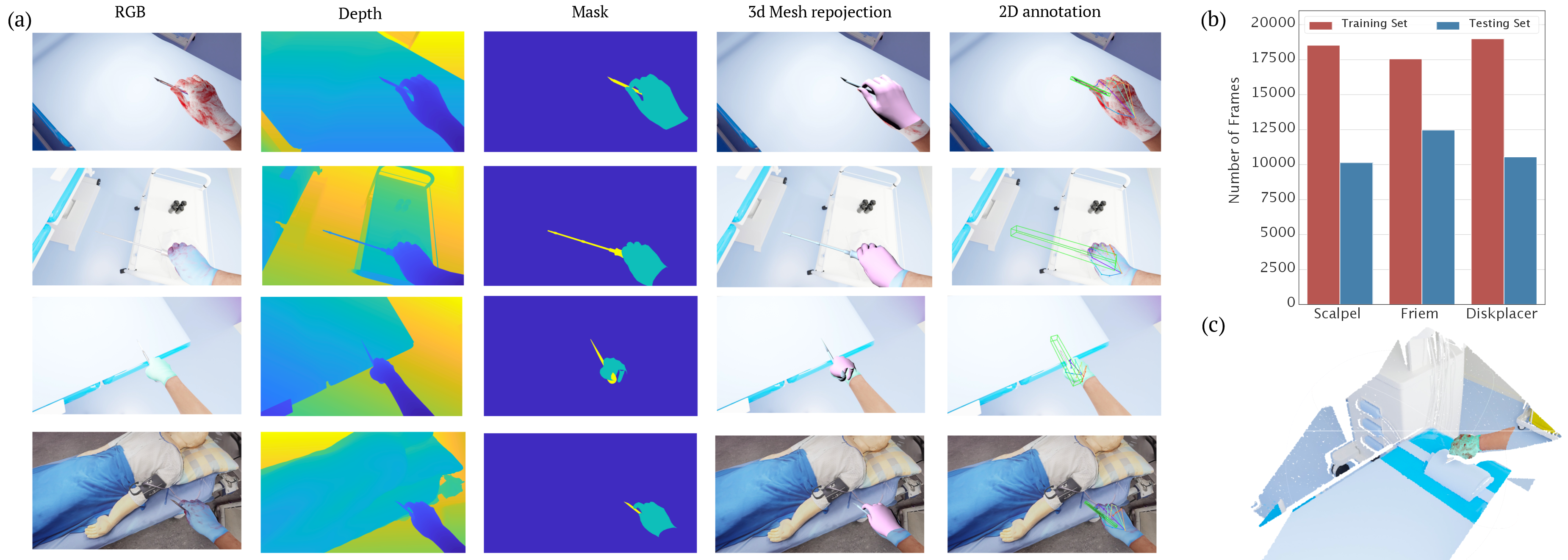}
    \caption{(a) Dataset samples for RGB-D sequences and annotation. An example of the scalpel, friem, and diskplacer, is shown in the first three rows. The fourth row shows an example of the new scene and blood glove patterns that only appear in the test set. (b) shows the statistics on the number of frames for each surgical instrument in the training and testing sets. (c) shows a point cloud created from an RGB-D frame with simulated Kinect noise.}
    \label{f:samples}
\end{figure}
We use blender \cite{blender2018} and bpycv packages to render the RGB-D sequences and instance segmentation masks. Diverse textures and scenes of high quality are provided in the dataset: it includes 24 SMPLX textures featuring blue, green, and white surgical gloves textures with various blood patterns and a synthetic surgical room scene created by artists. The Cycle rendering engine and de-noising post-processing are adopted to produce high-quality frames. POV-Surgery provides clean depth maps for depth-based methods or point-cloud-based methods, as the artifact of real depth cameras can be efficiently simulated via previous works as \cite{handa:etal:2014}. A point cloud example generated from an RGB-D frame with added simulated Kinect depth camera noise is provided in Fig~\textbf{\ref{f:samples}}. The POV-Surgery dataset consists of 36 sequences with 55,078 frames in the training set and 17 sequences with 33,161 frames in the testing set, respectively. Three bloodied glove textures and one scene created from a room scanning of a surgery room are only used in the testing set to measure generalizability. Fig~\textbf{\ref{f:samples}} shows the ground truth data samples and the dataset statistics.
\section{Experiment}
\begin{figure}[htbp]
    \centering
    \includegraphics[width=0.8\linewidth]{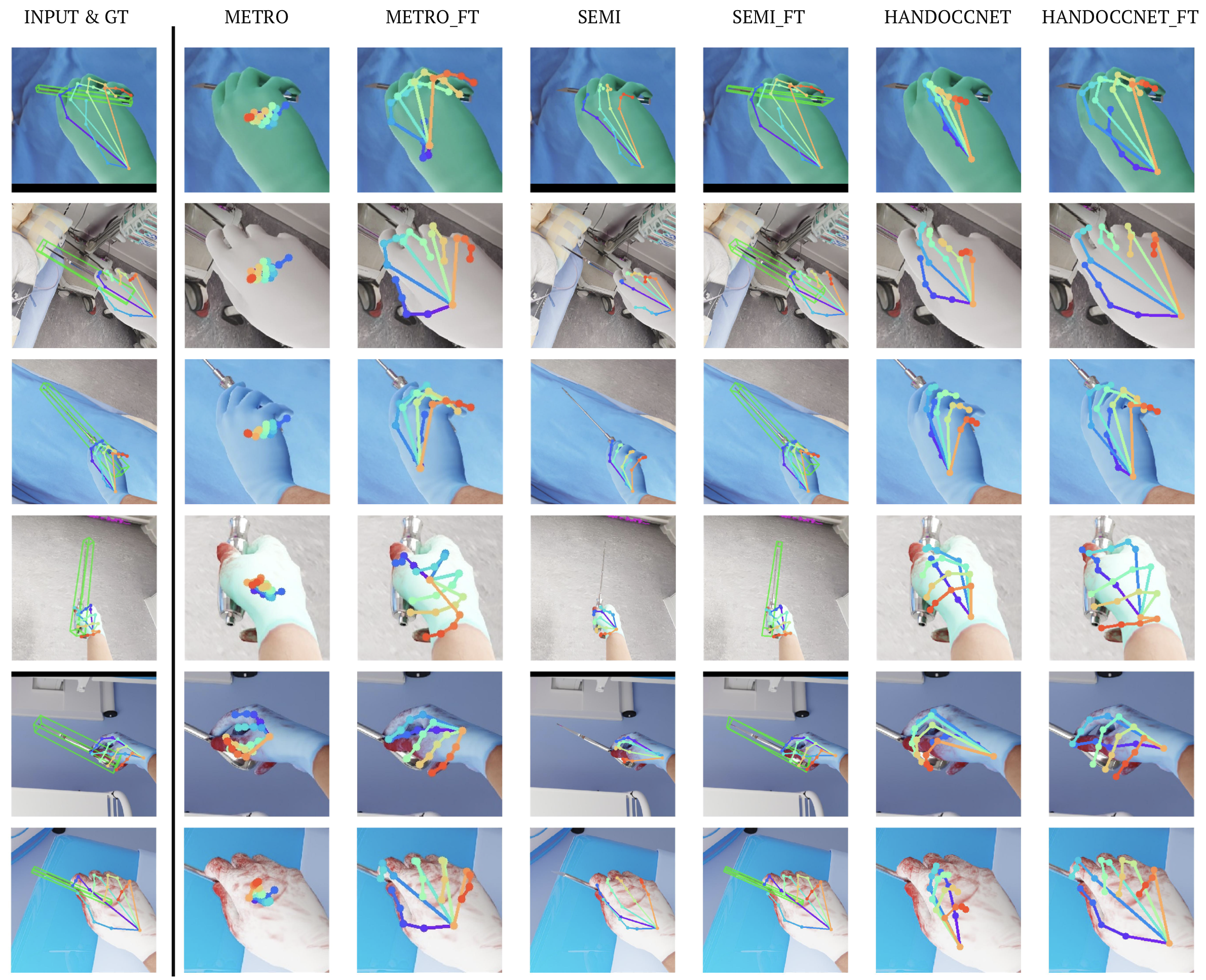}
    \caption{Qualitative results of METRO\cite{lin2021end-to-end}, SEMI\cite{liu2021semi}, and HANDOCCNET\cite{Park_2022_CVPR_HandOccNet} on the test of of POV-Surgery. The FT denotes fine-tuning. We show the 2D re-projection of the predicted 3D hand joints and object control bounding box overlayed on the input image.}
    \label{f:quali}
\end{figure}
\begin{table}[t]
    \centering
\caption{The evaluation result of different methods on the test set of POV-Surgery, where the $_{ft}$ denotes fine-tuned on the training set. $P_{2d}$ denotes the 2D hand joint re-projection error (in pixels). MPJPE and PVE denote the 3D Mean Per Joint Position Error and Per Vertex Error, respectively. PA denotes procrustes alignment.}\label{tab1}
\begin{tabular}{cccccc}
\hline
Method \quad &  $P_{2d}\downarrow$ \quad & MPJPE $\downarrow$\quad & PVE $\downarrow$ &  PA-MPJPE $\downarrow$ &  PA-PVE $\downarrow$\\
\hline
METRO\cite{lin2021end-to-end} & 95.11 & 77.46 & 75.06 & 23.43 & 22.34 \\
SEMI\cite{liu2021semi} & 77.91 & 115.67 & 112.10 & 12.68 & 12.76 \\
HandOCCNet\cite{Park_2022_CVPR_HandOccNet} & 64.70 & 95.19 & 90.83 & 11.71 & 11.13 \\
METRO$_{ft}$ & 30.49 & 14.90 & 13.80 & 6.36 & 4.34 \\
SEMI$_{ft}$ & \textbf{13.42} & 15.14 & 14.69 & \textbf{4.29} & \textbf{4.23} \\
HandOCCNet$_{ft}$ & 13.80 & \textbf{14.35} & \textbf{13.73} & 4.49 & 4.35 \\
\hline
\end{tabular}
\end{table}
We evaluate and fine-tune two state-of-the-art hand pose estimation methods: \cite{lin2021end-to-end,Park_2022_CVPR_HandOccNet},  and one hand-object pose estimation\cite{liu2021semi} method on our dataset with provided checkpoints in their official repositories. 6 out of 36 sequences from the training set are selected as the validation set for model selection. We continue to train their checkpoints on our synthetic training set, with a reduced learning rate (10$^{-5}$) and various data augmentation methods such as color jittering, scale and center jittering, hue-saturation-contrast value jittering, and motion blur for better generalizability. Afterwards, we evaluate the performance of those methods on our testing set. We set a baseline for object control point error in pixels: 41.56 from fine-tuning \cite{liu2021semi}.  The hand quantitative metrics are shown in Tab.~\textbf{\ref{tab1}} and qualitative visualizations are shown in Fig.~\textbf{\ref{f:quali}}, where we highlight the significant performance improvement for existing methods after fine-tuning them on the POV-Surgery dataset.

\begin{figure}[htbp]
    \centering
    \includegraphics[width=\linewidth]{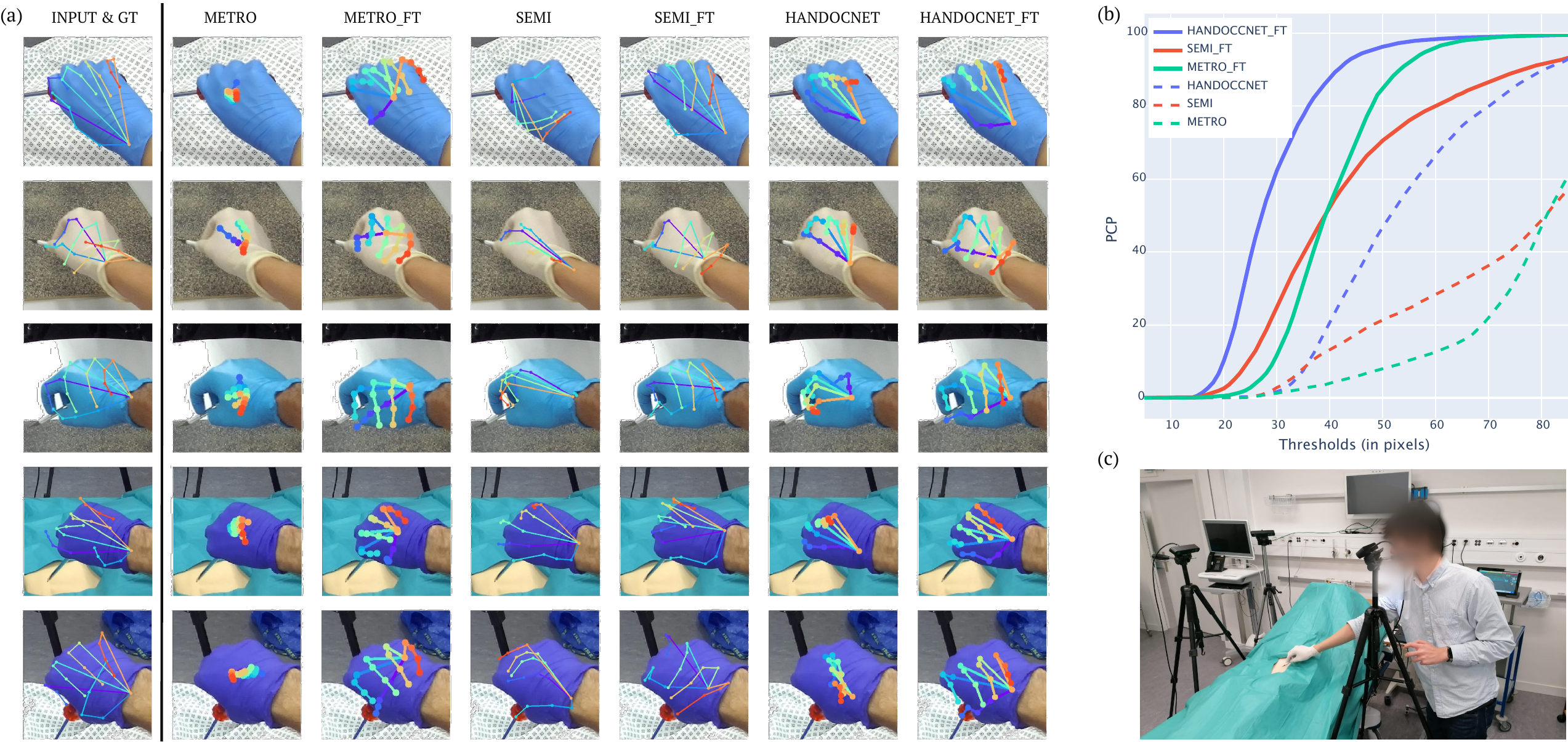}
    \caption{(a) Ground truth and qualitative results of different methods on the \textit{real-life test set}. (b) Accuracy with different 2D pixel error thresholds, showing large performance improvement after fine-tuning on POV-Surgery (c) Our multi-camera real-life data capturing set-up.}
    \label{f:Reallife}
\end{figure}

To further evaluate the generalizability of the methods fine-tuned on our dataset, we collect 6,557 real-life images with multiple surgical gloves, tools, and backgrounds as the \textit{real-life test set}. The data capture setup with four stereo cameras is shown in Fig~\textbf{\ref{f:Reallife}}. We adopt a top-down-based method from \cite{mmpose2020} with manually selected hand bounding boxes for 2D hand joint detection. \cite{easymocap} is used to reconstruct 3D hand poses from different camera observations. We project the hand pose to the egocentric camera view and manually select the frames with accurate hand predictions to obtain reliable 2D hand pose ground truth. We show quantitative examples of the indicated methods and the PCP curve in Fig~\textbf{\ref{f:Reallife}}. After fine-tuning on our synthetic dataset significant performance improvements are achieved for SOTA methods on the \textit{real-life test set}. Particularly, we observe a similar performance improvement for unseen purple-texture gloves, showing the effectiveness of our POV-Surgery dataset towards the challenging egocentric surgical hand-object interaction scenarios in general.
\section{Conclusion}
This paper proposes a novel synthetic data generation pipeline that generates hand-tool manipulation temporal sequences. Using the data generation pipeline and focusing on three tools used in orthopedic surgeries: scalpel, diskplacer, and friem,  we propose a large, synthetic, and temporal dataset on egocentric surgical hand-object pose estimation, with 88,329 RGB-D frames and diverse bloody surgical gloves patterns. We evaluate and fine-tune three current state-of-the-art methods on the POV-Surgery dataset. We prove the effectiveness of the synthetic dataset by showing the significant performance improvement of the SOTA methods in real-life cases with surgical gloves and tools.
\section{Acknowledgement}
This work is part of a research project that has been financially supported by Accenture LLP. Siwei Zhang is funded by Microsoft Mixed Reality \& AI Zurich Lab PhD scholarship. The authors would like to thank PD Dr. Michaela Kolbe for providing the simulation facilities and the students participating in motion capture.

\bibliographystyle{splncs04}
%
\bibliography{refs}

\end{document}